\pdfoutput=1
\documentclass[11pt,a4paper]{article}
\usepackage[hyperref]{acl2018}
\usepackage{times}
\usepackage{latexsym}
\usepackage{url}
\usepackage{array}
\usepackage{graphicx} 
\usepackage{amsmath,amsfonts,amssymb}

\makeatletter
\DeclareRobustCommand{\cev}[1]{%
	\mathpalette\do@cev{#1}%
}
\newcommand{\do@cev}[2]{%
	\fix@cev{#1}{+}%
	\reflectbox{$\m@th#1\vec{\reflectbox{$\fix@cev{#1}{-}\m@th#1#2\fix@cev{#1}{+}$}}$}%
	\fix@cev{#1}{-}%
}
\newcommand{\fix@cev}[2]{%
	\ifx#1\displaystyle
	\mkern#23mu
	\else
	\ifx#1\textstyle
	\mkern#23mu
	\else
	\ifx#1\scriptstyle
	\mkern#22mu
	\else
	\mkern#22mu
	\fi
	\fi
	\fi
}

\aclfinalcopy 

\title{Learning to Automatically Generate Fill-In-The-Blank Quizzes}

\author{Edison Marrese-Taylor, Ai Nakajima, Yutaka Matsuo\\
        Graduate School of Engineering\\
        The University of Tokyo\\
        {\tt emarrese,ainakajima,matsuo@weblab.t.u-tokyo.ac.jp}\\\AND
        Ono Yuichi\\
        Center for Education of Global Communication\\
        University of Tsukuba\\
        {\tt ono.yuichi.ga@u.tsukuba.ac.jp}\\}

\date{}

\begin{document}

\maketitle
\begin{abstract}
In this paper we formalize the problem automatic fill-in-the-blank question generation using two standard NLP machine learning schemes, proposing concrete deep learning models for each. We present an empirical study based on data obtained from a language learning platform showing that both of our proposed settings offer promising results.
\end{abstract}

\section{Introduction}

With the advent of the Web 2.0, regular users were able to share, remix and distribute content very easily. As a result of this process, the Web became a rich interconnected set of heterogeneous data sources. Being in a standard format, it is suitable for many tasks involving knowledge extraction and representation. For example, efforts have been made to design games with the purpose of semi-automating a wide range of knowledge transfer tasks, such as educational quizzes, by leveraging on this kind of data.

In particular, quizzes based on multiple choice questions (MCQs) have been proved efficient to judge students’ knowledge. However, manual construction of such questions often results a time-consuming and labor-intensive task. 

Fill-in-the-blank questions, where a sentence is given with one or more blanks in it, either with or without alternatives to fill in those blanks, have gained research attention recently. In this kind of question, as opposed to MCQs, there is no need to generate a WH style question derived from text. This means that the target sentence could simply be picked from a document on a corresponding topic of interest which results easier to automate.

Fill-in-the-blank questions in its multiple-choice answer version, often referred to as cloze questions (CQ), are commonly used for evaluating proficiency of language learners, including official tests such as TOEIC and TOEFL \cite{sakaguchi_discriminative_2013}. They have also been used to test students knowledge of English in using the correct verbs \cite{sumita_measuring_2005}, prepositions \cite{lee_automatic_2007} and adjectives \cite{lin_automatic_2007}. \citet{pino_selection_2008} and \citet{smith_gap-fill_2010} generated questions to evaluate student’s vocabulary. 

The main problem in CQ generation is that it is generally not easy to come up with appropriate distractors ---incorrect options--- without rich experience. Existing approaches are mostly based on domain-specific templates, whose elaboration relies on experts. Lately, approaches based on discriminative methods, which rely on annotated training data, have also appeared. Ultimately, these settings prevent end-users from participating in the elaboration process, limiting the diversity and variation of quizzes that the system may offer. 

In this work we formalize the problem of automatic fill-in-the-blank question generation and present an empirical study using deep learning models for it in the context of language learning. Our study is based on data obtained from our language learning platform \cite{nakajima2013new,ono_nakajima_2015,ono2017motivational} where users can create their own quizzes by utilizing freely available and open-licensed video content on the Web. In the platform, the automatic quiz creation currently relies on hand-crafted features and rules, making the process difficult to adapt. Our goal is to effectively provide an adaptive learning experience in terms of style and difficulty, and thus better serve users' needs \cite{lin_sherlock:_2015}. In this context, we study the ability of our proposed architectures in learning to generate quizzes based on data derived of the interaction of users with the platform. 

\section{Related Work}

The problem of fill-in-the-blank question generation has been studied in the past by several authors. Perhaps the earlies approach is by \citet{sumita_measuring_2005}, who proposed a cloze question generation system which focuses on distractor generation using search engines to automatically measure English proficiency. In the same research line, we also find the work of \citet{lee_automatic_2007}, \citet{lin_automatic_2007} and \citet{pino_selection_2008}. In this context, the work of \citet{goto_automatic_2009} probably represents the first effort in applying machine learning techniques for multiple-choice cloze question generation. The authors propose an approach that uses conditional random fields \cite{lafferty_conditional_2001} based on hand-crafted features such as word POS tags. 

More recent approaches also focus on the problem of distractor selection or generation but apply it to different domains. For example, \citet{narendra_automatic_2013}, present a system which adopts a semi-structured approach to generate CQs by making use of a knowledge base extracted from a Cricket portal. On the other hand, \citet{lin_sherlock:_2015} present a generic semi-automatic system for quiz generation using linked data and textual descriptions of RDF resources. The system seems to be the first that can be controlled by difficulty level. Authors  tested it using an on-line dataset about wildlife provided by the BBC. \citet{kumar_automatic_2015} present an approach automatic for CQs generation for student self-assessment.

Finally, the work of \citet{sakaguchi_discriminative_2013} presents a discriminative approach based on SVM classifiers for distractor generation and selection using a large-scale language learners’ corpus. The SVM classifier works at the word level and takes a sentence in which the target word appears, choosing a verb as the best distractor given the context. Again, the SVM is based on human-engineered features such as n-grams, lemmas and dependency tags.

Compared to approaches above, our take is different since we work on fill-in-the-blank question generation without multiple-choice answers. Therefore, our problem focuses on word selection ---the word to blank--- given a sentence, rather than on distractor generation. To the best of our knowledge, our system is also the first to use representation learning for this task.

\section{Proposed Approach}
 
We formalize the problem of automatic fill-on-the-blanks quiz generation using two different perspectives. These are designed to match with specific machine learning schemes that are well-defined in the literature. In both cases. we consider a training corpus of $N$ pairs $(S_n, C_n), \: n = 1 \ldots N$ where $S_n = s_1, \ldots, s_{L(S_n)}$ is a sequence of $L(S_n)$ tokens and $C_n \in  [1, L(S_n)]$ is an index that indicates the position that should be blanked inside $S_n$.

This setting allows us to train from examples of single blank-annotated sentences. In this way, in order to obtain a sentence with several blanks, multiple passes over the model are required. This approach works in a way analogous to humans, where blanks are provided one at a time.

\subsection{AQG as Sequence Labeling}
 
Firstly, we model the AQG as a sequence labeling problem. Formally, for an embedded input sequence $S_n = s_1, \ldots, s_{L(n)}$ we build the corresponding label sequence by simply creating a one-hot vector of size $L(S_n)$ for the given class $C_n$. This vector can be seen as a sequence of binary classes, $Y_n =y_1, \ldots, y_{L(n)}$, where only one item (the one in position $C_n$) belongs to the positive class. Given this setting, the conditional probability of an output label is modeled as follows:
\begin{eqnarray}
p(y \mid s) \propto \prod_{i=1}^{n} \hat y_i \\
\hat y_i = H(y_{i-1},y_i, s_i) 
\end{eqnarray}
Where, in our, case, function $H$ is modeled using a bidirectional LSTM \cite{hochreiter1997long}. Each predicted label distribution $\hat y_t$ is then calculated using the following formulas. 
\begin{eqnarray}
\vec h_i = LSTM_{fw}(\vec h_{i-1}, x_i)\\
\cev h_i = LSTM_{bw}(\cev h_{i+1}, x_i)\\
\hat y_i = \mathrm{softmax} ( [ \vec h_i ; \cev{h_i} ])
\end{eqnarray}

\begin{figure}[h!]
	\centering
    \includegraphics[width=0.6\columnwidth]{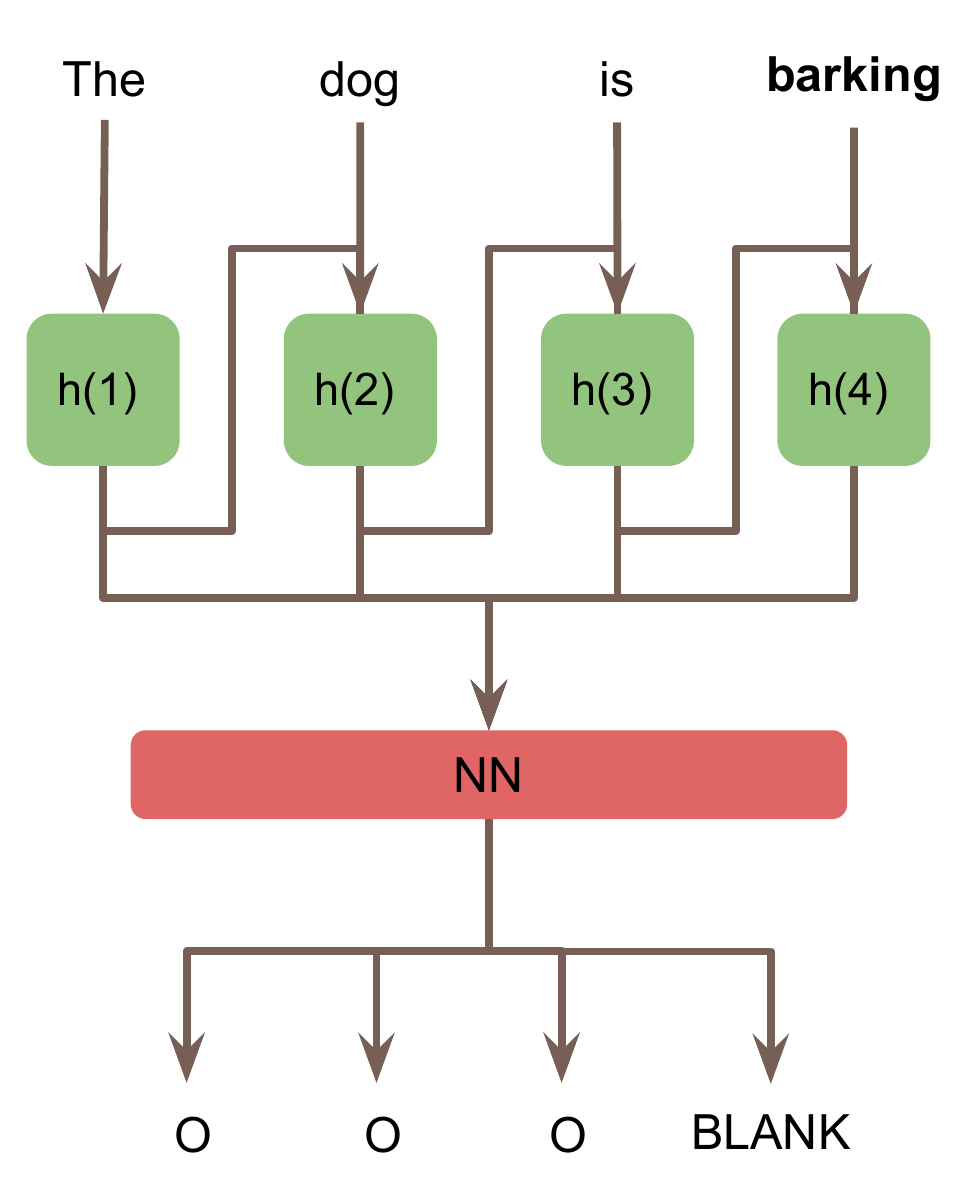}
	\caption{Our sequence labeling model based on an LSTM for AQG.}
	\label{fig:bilstm}
\end{figure}

The loss function is the average cross entropy for the mini-batch. Figure \ref{fig:bilstm} summarizes the proposed model.

\begin{equation}
L(\mathbf{\theta})\ = -\frac1n\sum_{i=1}^n\ y_i \log \hat y_i + (1 - y_i)  \log (1 - \hat y_i)
\end{equation}

 \subsection{AQG as Sequence Classification}

\begin{figure}[h!]
	\centering
    \includegraphics[width=0.9\columnwidth]{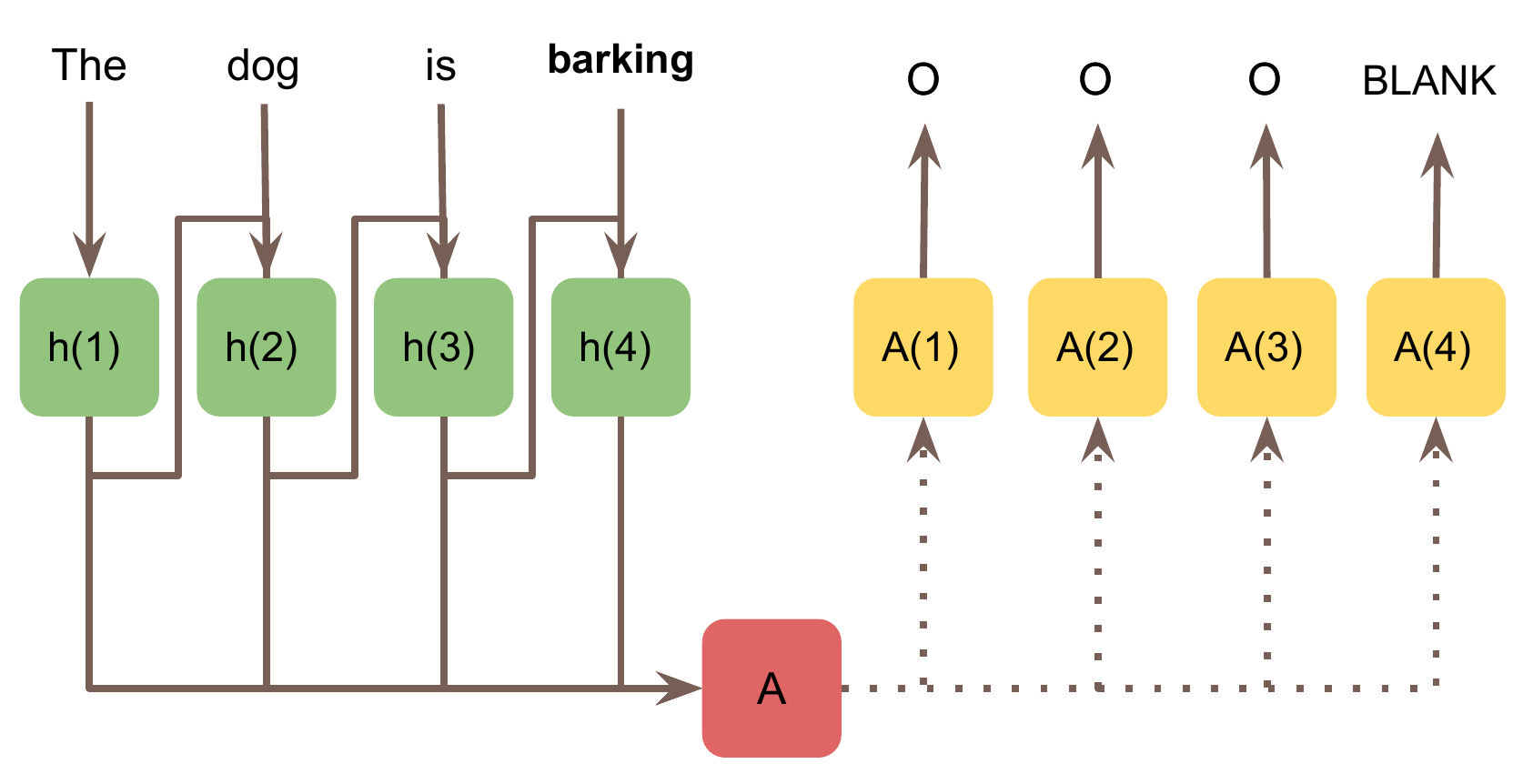}
	\caption{Our sequence classification model, based on an LSTM for AQG.}
	\label{fig:prtnet}
\end{figure}

In this case, since the output of the model is a position in the input sequence $S_n$, the size of output dictionary for $C_n $ is variable and depends on $S_n$. Regular sequence classification models use a softmax distribution over a fixed output dictionary to compute $p(C_n|S_n$) and therefore are not suitable for our case. Therefore, we propose to use an attention-based approach that allows us to have a variable size dictionary for the output softmax, in a way akin to Pointer Networks \cite{vinyals_pointer_2015}. More formally, given an embedded input vector sequence $S_n = s_1, ..., s_{L(n)}$, we use a bidirectional LSTM to first obtain a dense representation of each input token.
\begin{eqnarray}
\vec h_i = LSTM_{fw}(\vec h_{i-1}, x_i)\\
\cev h_i = \cev LSTM_{bw}(\cev h_{i+1}, x_i)\\
h_i = [\vec h_i; \cev h_i]
\end{eqnarray}
We later use pooling techniques including $max$ and $mean$ to obtain a summarized representation $\bar h$ of the input sequence, or simply take the $last$ hidden state as a drop-in replacement to do so. After this, we add a global content-based attention layer, which we use to to compare that summarized vector to each hidden state $h_i$. Concretely,
\begin{eqnarray}
u = v^{\intercal} W[h_i ; \bar h ]\\
p(C_n|P_n)=softmax(u)
\end{eqnarray}
Where $W$ and $v$ are learnable parameters of the model, and the softmax normalizes the vector $u$ to be an output distribution over a dictionary of size $L(S_n)$. Figure \ref{fig:prtnet} summarizes the proposed model graphically. Then, for a given sentence $C_k$, the goal of our model is to predict the most likely position $C^{\star} \in [1, L(S_n)]$ of the next word to be blanked.

\section{Empirical Study}

Although the hand-crafted rule-based system currently used in our language learning platform offers us good results in general, we are interested in developing a more flexible approach that is easier to tailor depending on the case. In particular, in an adaptive learning setting where the goal is resource allocation according to the unique needs of each learner, rule-based methods for AQG appear to have insufficient flexibility and adaptability to accurately model the features of each learner or teacher.

With this point in mind, this section presents an empirical study using state-of-the-art Deep Learning approaches for the problem of AQG. In particular, the objective is to test to what extent our prosed models are able to encode the behavior of the rule-based system. Ultimately, we hope that these can be used for a smooth transition from the current human-engineered feature-based system to a fully user-experience-based regime. 

In Natural Language Processing, deep models have succeeded in large part because they learn and use their own continuous numeric representational systems for words and sentences. In particular, distributed representations \cite{hinton1984distributed} applied to words \cite{mikolov_distributed_2013} have meant a major breakthrough. All our models start with random word embeddings, we leave the usage of other pre-trained vectors for future work.

Using our platform, we extracted anonymized user interaction data in the manner of real quizzes generated for a collection of several input video sources. We obtained a corpus of approximately 300,000 sentences, from which roughly 1.5 million single-quiz question training examples were derived. We split this dataset using the regular 70/10/20 partition for training, validation and testing.

As the system required the input sentences to be tokenized and makes use of features such as word pos-tags and such, the sentences in our dataset are processed using CoreNLP \cite{manning-EtAl:2014:P14-5}. We also extract user-specific and quiz-specific information, including word-level learning records of the user, such as the number of times the learner made a mistake on that word, or whether the learner looked up the word in the dictionary. In this study, however, we restrain our model to only look at word embeddings as input.

We use the same data pre-processing for all of our models. We build the vocabulary using the train partition of our dataset with a minimum frequency of 1. We do not keep cases and obtain an unknown vocabulary of size 2,029, and a total vocabulary size of 66,431 tokens.

\subsection{Sequence Labeling}

We use a 2-layer bidirectional LSTM, which we train using Adam \citet{DBLP:journals/corr/KingmaB14} with a learning rate of $0.001$, clipping the gradient of our parameters to a maximum norm of 5. We use a word embedding size and hidden state size of 300 and add dropout \cite{srivastava2014dropout} before and after the LSTM, using a drop probability of 0.2. We train our model for up to 10 epochs. Training lasts for about 3 hours. 

For evaluation, as accuracy would be extremely unbalanced given the nature of the blanking scheme ---there is only one positive-class example on each sentence--- we use Precision, Recall and F1-Score over the positive class for development and evaluation. Table \ref{table:results_labeling} summarizes our obtained results.
\begin{table}[h!]
  \centering
  \begin{tabular}{c | c | c | c | c}
  \textbf{Set} & \textbf{Loss} & \textbf{Prec.} & \textbf{Recall} & \textbf{F1-Score} \\
  \hline
  Valid  & 0.0037 & 88.35 & 88.81 & 88.58 \\
  Test  & 0.0037 & 88.56 & 88.34 & 88.80
  \end{tabular}
  \caption{Results  of the seq. labeling approach.}
  \label{table:results_labeling}
\end{table}

\subsection{Sequence Classification}

In this case, we again use use a 2-layer bidirectional LSTM, which we train using Adam with a learning rate of $0.001$, also clipping the gradient of our parameters to a maximum norm of 5. Even with these limits, convergence is faster than in the previous model, so we only trained the the classifier for up to 5 epochs. Again we use a word embedding and hidden state of 300, and add dropout with drop probability of 0.2 before and after the LSTM. Our results for different pooling strategies showed no noticeable performance difference in preliminary experiments, so we report results using the last hidden state.

For development and evaluation we used accuracy over the validation and test set, respectively. Table \ref{table:results_classification} below summarizes our obtained result, we can see that model was able to obtain a maximum accuracy of approximately 89\% on the validation and testing sets.
\begin{table}[h!]
  \centering
  \begin{tabular}{c | c | c }
  \textbf{Set} & \textbf{Loss} & \textbf{Accuracy}\\
  \hline
  Valid  & 101.80 & 89.17 \\
  Test  & 102.30 & 89.31 
  \end{tabular}
  \caption{Results of the seq. classification approach.}
  \label{table:results_classification}
\end{table}

\section{Conclusions}

In this paper we have formalized the problem of automatic fill-on-the-blanks quiz generation using two well-defined learning schemes: sequence classification and sequence labeling. We have also proposed concrete architectures based on LSTMs to tackle the problem in both cases.

We have presented an empirical study in which we test the proposed architectures in the context of a language learning platform. Our results show that both the0 proposed training schemes seem to offer fairly good results, with an Accuracy/F1-score of nearly 90\%. We think this sets a clear future research direction, showing that it is possible to transition from a heavily hand-crafted approach for AQG to a learning-based approach on the base of examples derived from the platform on unlabeled data. This is specially important in the context of adaptive learning, where the goal is to effectively provide an tailored and flexible experience in terms of style and difficulty

For future work, we would like to use different pre-trained word embeddings as well as other features derived from the input sentence to further improve our results. We would also like to test the power of the models in capturing different quiz styles from real questions created by professors.

\bibliographystyle{acl_natbib}
\bibliography{acl2018}

\end{document}